\documentclass[runningheads]{llncs}
\usepackage{graphicx}
\usepackage{amsmath,amssymb} 
\usepackage{color}

\begin{document}
\pagestyle{headings}
\mainmatter

\def\ACCV18SubNumber{938}  

\title{Tournament Based Ranking CNN \\for the  Cataract grading} 
\titlerunning{Tournament based Ranking CNN for the Cataract grading}
\authorrunning{D. Kim et al.}

\author{Dohyeun Kim\inst{1}, Tae Joon Jun\inst{1}, Daeyoung Kim\inst{1}, Youngsub Eom\inst{2}} 
\institute{KAIST, Daejeon, Republic of Korea\\
\and
Korea University College of Medicine, Seoul, Republic of Korea\\
}

\maketitle

\begin{abstract}
Solving the classification problem, unbalanced number of dataset among the classes often causes performance degradation. Especially when some classes dominate the other classes with its large number of datasets, trained model shows low performance in identifying the dominated classes. This is common case when it comes to medical dataset. Because the case with a serious degree is not quite usual, there are imbalance in number of dataset between severe case and normal cases of diseases. Also, there is difficulty in precisely identifying grade of medical data  because of vagueness between them. To solve these problems, we propose new architecture of convolutional neural network named {\bf Tournament based Ranking CNN} which shows remarkable performance gain in identifying dominated classes while trading off very small accuracy loss in dominating classes. Our Approach complemented problems that occur when method of Ranking CNN that aggregates outputs of multiple binary neural network models is applied to medical data. By having tournament structure in aggregating method and using very deep pretrained binary models, our proposed model recorded 68.36\% of exact match accuracy, while Ranking CNN recorded 53.40\%, pretrained Resnet recorded 56.12\% and CNN with linear regression recorded 57.48\%. As a result, our proposed method is applied efficiently to cataract grading which have ordinal labels with imbalanced number of data among classes, also can be applied further to medical problems which have similar features to cataract and similar dataset configuration.  
\end{abstract}

\section{Introduction}

Cataract is one of the most frequent cause for blindness that crystalline lens become cloudy and having blurred vision as a consequence. The cloudy progress often proceeds slowly, finally leads patient to blindness. Because it can cause blindness when it is exacerbated, early discovery of cataract and continuous observation of crystalline lens state is necessary. Cataract progress can be expressed into Nuclear Cataract Grade. This Nuclear Cataract Grade has 6 grade 1 to 6, 1 is the mildest and 6 is the severe cataract grade. This grading is decided through slit lamp image. Slit lamp image is acquired from Slit lamp microscope, which is usually used for observing front side of eyeball including cornea, crystalline lens and Iris. Figure 1 shows Slit lamp image with 6 Nuclear Cataract Grades. Based on lens state and grade, doctors can diagnosis whether patients need operation or not. Grade 1 and 2 are regarded as Normal but patient who has over grade 3 should consider Cataract surgery in order to cure blurred eye sight and blindness. Thus classifying cataract grade is important to early finding of cataract and preventing it. 

To aid Cataract grading there has been several approaches in automation of cataract grading. Many of them extracted feature from slit-lamp image and applied classifier to grade cataract\cite{nayak2013automated,huang2009computer,li2010computer,xu2013automatic}. Jagadish Nayak classified Cataract using SVM classifier with detecting big ring-area of eye and extracting features such as edge pixel count\cite{nayak2013automated}. Huiqi Li et al. proposed effective method of analyzing structure of lens by modeling lens structure and finding matching point between image and shape model. Also they classified cataract degree with SVM regressors after extracting feature such as intensity\cite{li2010computer}. Yanwu Xu et al. extracted feature by detecting structure of eye lens using method of Li et al.\cite{li2010computer}, then used group sparsity Regression\cite{xu2013automatic}.  Wei Huang et al. suggested ranking function which sorts group of slit-lamp image according to severity of cataract\cite{huang2009computer} while using lens structure method of \cite{li2010computer}. After Deep Learning showed impressive performance in various fields, several approach used neural network to grade cataract. Xiyang Liu et al. proposed cataract diagnosis framework using convolutional neural network with SVM classifier while Localizing ROI(region of Interest) using Hough transform algorithm\cite{liu2017localization}. Xinting Gao et al. proposed method of Deep learning to obtain high level feature then using svm regressor to grade cataract\cite{gao2015automatic}. Yang et al. used ensemble learning for grading cataract with fundus images\cite{yang2016exploiting}

But these approaches \cite{nayak2013automated,huang2009computer,li2010computer,xu2013automatic,liu2017localization,gao2015automatic,yang2016exploiting}didn't consider characteristic of cataract and medical dataset which can cause degradation of performance. While it is easy to acquire normal case data, getting data of severe illness case is very hard because number of patients are relatively small compared to those who don't have diseases. Consequently dataset consists of a lots of normal data, but small number of extreme case data such as having severe disease. Such imbalance between normal case and disease case hinder proper learning to machine learning model, leading the model to be shifted to more prevalent data. For example, Cataract grade 5,6 and 1 is very rare because they are extreme case of both severe cataract and normal. When the model is trained without any proper measure, the trained model is likely to predict input image to be prevalent case without capturing distinctive feature of it because the model learned more frequently dominating cases. As a result, this leads high accuracy of dominating classes but low sensitivity to dominated classes. By means of preventing illness, acquiring high sensitivity of predicting disease case is important.
Also, ambiguity between adjacent class hinder acquiring precise prediction. In figure 1, training model to identify images in class 5 and 6 is very complicated task. However at some point, such as classifying class 4 and 5, the task is much easier because opacity increases significantly. This means label and grade of cataract is not in exact linear relation. 

To handle these problems, we introduce tournament based convolutional neural network, which has effect of balancing biased number of training data among classes, and handling ambiguity within adjacent classes by having tournament structure in classification process. Our method Tournament based Ranking CNN consists of tournament structure and binary CNN models. Tournament structure divides class set into two subset recursively until each subset has only one class label. Binary CNN models classifies two subsets that is divided by tournament structure. Class of input image is predicted when input image is finally classified into one subset which has only one class while consistently classified into one of two subsets by binary models. In this manner, tournament structure has effect of balancing number of images among classes, preventing biased learning toward dominating classes. 

We proposed 3 ways of consisting tournament structure, dividing according to best AUC, balancing number of images in two subsets, and balancing number of classes in two subsets. AUC based tournament structure recorded best accuracy of 68.36\%, showing impressive result in classifying class 1 and 6. AUC Based Tournament model recorded 50\%of accuracy in class 1 and 6, but CNN with linear model and Ranking CNN model failed classification as they predicted all class 1 and 6 images as class 2 and 5. Resnet multi label classification model recorded 12\% and 50\% in class 1 and 6. Only in Class 3, CNN with linear regression model has better accuracy of 80\% compared to 73\% of AUC based tournament model because CNN with linear regression is trained biased to dominating class of 3. By dividing class into two subset in a way that recorded highest classification AUC, Tournament Based Ranking CNN has effect of balancing biased number of data among classes and preventing biased learning into dominating classes. Also, as it divided according to AUC, it alleviated complexness caused by ambiguity in adjacent classes and thus, achieved higher accuracy. Many medical data suffers from imbalance data because of scarcity of patients who have severe degrees of diseases. Also, like cataract, diseases which have grading systems suffers from ambiguity between grades. This work can be applied further to medical dataset which has these property, and non medical data that has similar characteristic. 

In the next section, we will introduce Ranking CNN and our Tournament based Ranking CNN model to address unbalanced class and vagueness problem in cataract dataset. Then experiment result with accuracy and visualization will be introduced in section 3, conclusion will be followed next.

{\centering
\begin{figure}
\centering
\begin{tabular}{c c c}
\includegraphics[height=3.5cm]{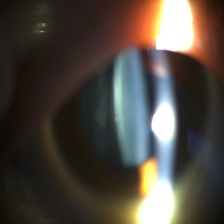}&\includegraphics[height=3.5cm]{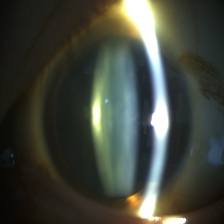}&\includegraphics[height=3.5cm]{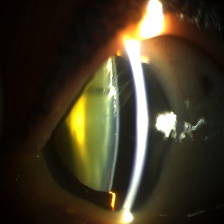}\\
Grade 1 &Grade 2 &Grade 3 \\

\includegraphics[height=3.5cm]{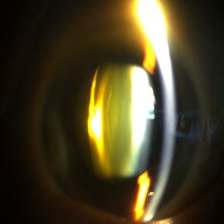} &
\includegraphics[height=3.5cm]{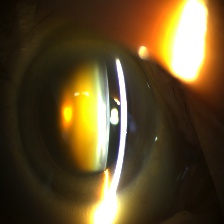} &
\includegraphics[height=3.5cm]{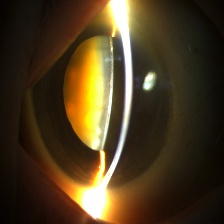}\\
Grade 4 &Grade 5 &Grade 6 \\
\end{tabular}
\caption{Slit-Lamp Images of Cataract. 
The image at the upper left corner is grade 1, and the lower right corner one is grade 6. As become closer to the grade 6, the opacity gets worse }
\end{figure}

}

\section{Methodology}

In this section we will introduce our Tournament based convolutional neural network model to deal with problem of unbalanced number of dataset, vagueness of adjacent classes and non linearity between label and cataract degree feature. Ranking CNN is proposed to solve ordinal dataset problem with non linearity pattern between label and features like cataract dataset. However, when applied to cataract dataset, it showed bad performance on classifying class 1 and 6 due to imbalance of data. So we briefly introduce Ranking CNN and then propose Tournament Based Ranking CNN which complement problems of Ranking CNN when applied to unbalanced dataset having ambiguity in adjacent classes. Our model consists of two steps. One is binary deep neural network, the other is constructing Tournament structure. 

\subsection{Ranking CNN}
To estimate age in human face image, Ranking CNN was proposed. Similar to Cataract grade, human age estimation dataset has ordinal property, which means there is an order between classes. However multi label classification often neglect these property because output neurons of individual classes calculate their result independently. As a result, output neuron can predict input image is in certain class, but having no idea of which one is older or younger. Neglecting this ordinal property hinder obtaining good prediction result. Moreover, aging process does not exactly follow linear relation with age, regression model can oversimplify to the linear model. To handle these problem,  Shin, Hoo-Chang, et al. proposed Ranking CNN\cite{chen2017using}. They trained multiple binary convolutional neural network models that predict input image is whether over or below the certain label. Suppose number of label is given K, K-1 number of models can be trained. In a way that aggregating result of K-1 binary models, achieving high accuracy with considering ordinal property was possible.  
However, Result of Ranking CNN on cataract dataset was not very successful. The reason is that grade 1 and 6 has high miss prediction rate due to small dataset and vagueness. As similarity between grade 1 and grade 2 is very high, when dataset is divided as grade 1 and the other, binary model can not successfully learn features that only distinguishes grade 1. Moreover, low number of training set on grade 1 also hinder achieving good accuracy. 
TJ Jun et al. proposed 2-stage Ranking CNN on glaucoma dataset\cite{jun20182sranking}. They applied Ranking CNN on glaucoma dataset which consists of 3 stage of glaucoma, Normal, Suspicious and Glaucoma. In 1st stage, 2 binary model classifies glaucoma and the others, and Normal and the others. After extracting ROI(Region of Interest) of image from Class Activation Map from previous model, ROI extracted images are trained by Ranking CNN same as 1st stage. Although they considered continuous and ambiguous problems in medical dataset, did not considered imbalance situation in medical datasets. 

\subsection{Tournament Based Ranking CNN}

To complement drawbacks of Ranking CNN due to unbalanced dataset, we propose tournament structure instead of simply aggregating binary result. Figure 2 describes architecture of Tournament based Ranking CNN. Tournament Ranking CNN consists of two part: Multiple binary convolution neural networks and tournament structure. Tournament structure divides ordered class sets into two subsets, images over class k or not. Each binary networks predicts which of subset that class of input image belongs to. Also, final convolution layers and weights of these binary networks are used for generating Class Activation Map, which visualizes evidence of classification results. As Tournament structure divides class sets until all of subsets only contains 1 class, final subset that input image belongs to become result class of prediction. 

{\centering
\begin{figure}[tbh!]
\centering
\includegraphics[width=0.9\textwidth]{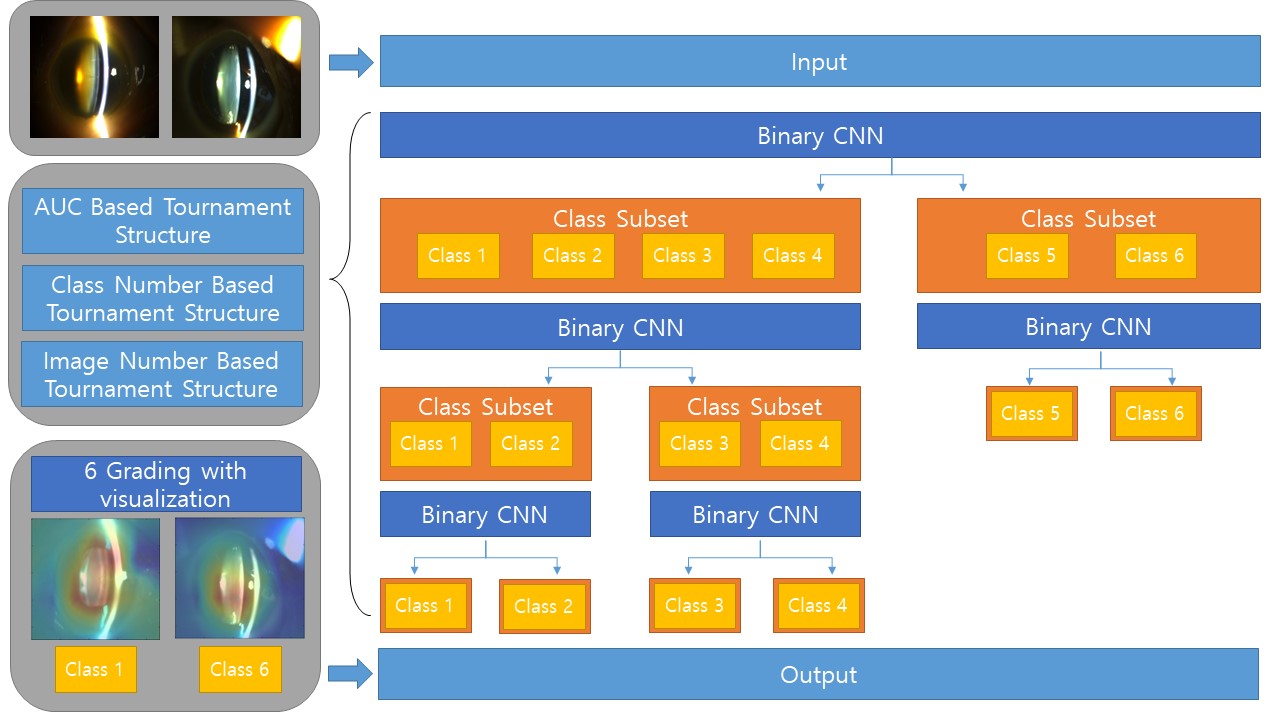}
\caption{Example structure of tournament based ranking cnn }
\end{figure}
}

To construct tournament structure, iteratively divide class label set into two subset until each subset consists only one class. Subset is divided by Criteria K that divides into two subsets that class is larger than K and equal to or less than K. Divided two subset become child of original subset. As a result tournament binary tree is constructed.

%
%
   

We propose 3 methods of choosing criteria K that decides structure of tree: method based on AUC of binary model classifies two subset, choosing criteria K that balances number of image between two subset, and that balances number of class between two subset. 

\subsubsection{Tournament structure based on AUC}
When choosing criteria K for dividing two subset, this method choose K based on AUC of binary model. Given N number of classes, there are N-1 number of possible way of dividing into two ordered set. For each case, train binary model to classify two subset and calculate AUC. The case where binary model recorded the highest AUC become criteria for dividing class set into two subset. According to criteria K, AUC differs a lot between cases. This is because there are some point K, opacity pattern of cataract changes so strongly compared to other point that classifying two subset become much easier. Also, model cannot find strong features that distinguish two subset when two similar grades are separated by criteria K. So we expected dividing subset according to AUC has effect of improved performance that separates ambiguous boundaries by distinguish clear boundaries first.

{\centering
\begin{figure}[tbh!]
\centering
\includegraphics[width=\textwidth]{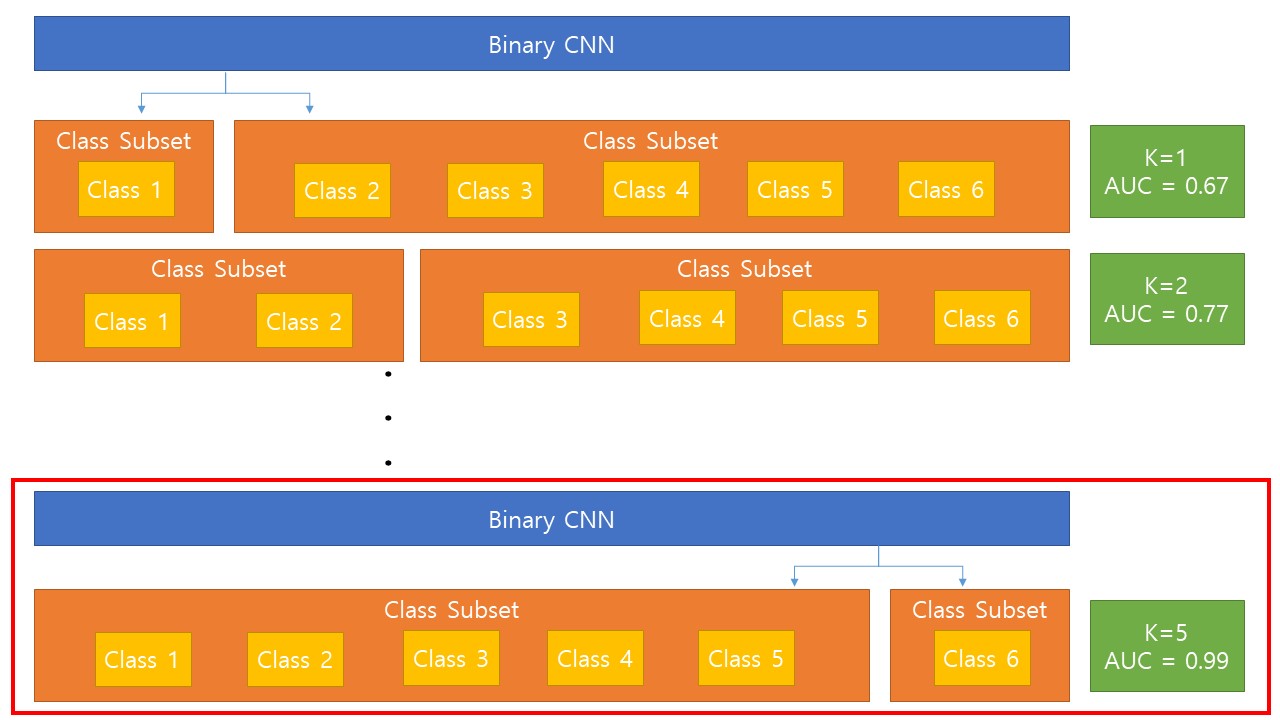}
\caption{Method of choosing K in AUC based tournament based Ranking CNN. Choose K that recorded highest AUC from possible dividing cases. }
\end{figure}
}

\subsubsection{Tournament structure based on balancing number of images}
This method choose K that balances number of images in divided subset. To balance number of image, this method select K that maximize (Number of image in subset 1) $\times$ (Number of image in subset 2). This is because (Number of image in subset 1) $\times$ (Number of image in subset 2) become maximized when the number of two is same each other. Reason for dividing in this way is that this method concentrates to alleviate biased learning toward dominating classes by grouping dominated classes together. Even one dominates other class by numerous data numbers, training process will not be biased to that class because classes are grouped in balanced way.

{\centering
\begin{figure}[tbh!]
\centering
\includegraphics[width=\textwidth]{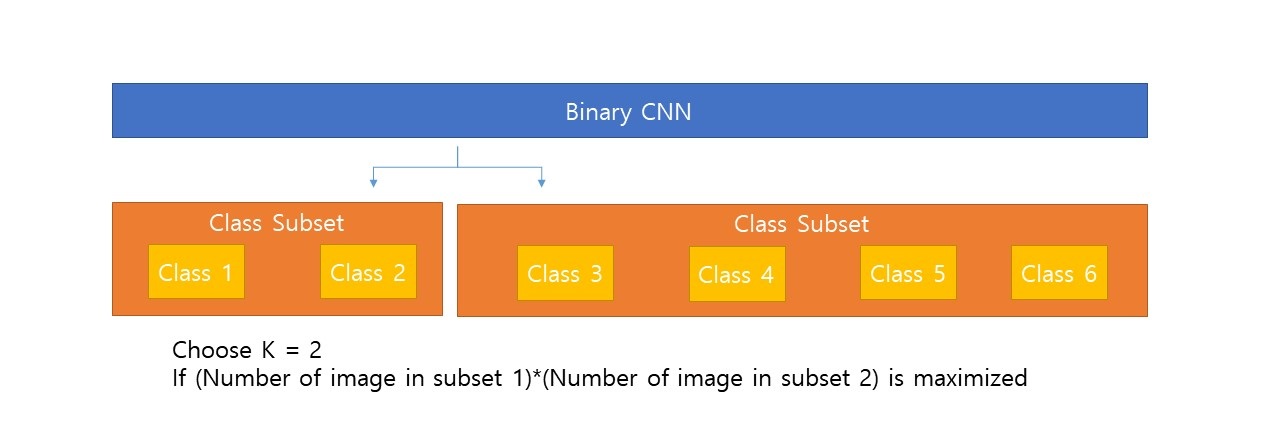}
\caption{Method of choosing K by balancing number of images. Choose K that balances number of images between two subsets the most. }
\end{figure}

}

\subsubsection{Tournament structure based on Balancing number of class}
This method choose K that balances number of class in divided subset. If number of class is even number, both subset will have half of number of classes. But in case that number of class is not even, one side of subset will have one more class if number of images in one side is smaller that the other side of subset. 

{\centering
\begin{figure}[tbh!]
\centering
\includegraphics[width=\textwidth]{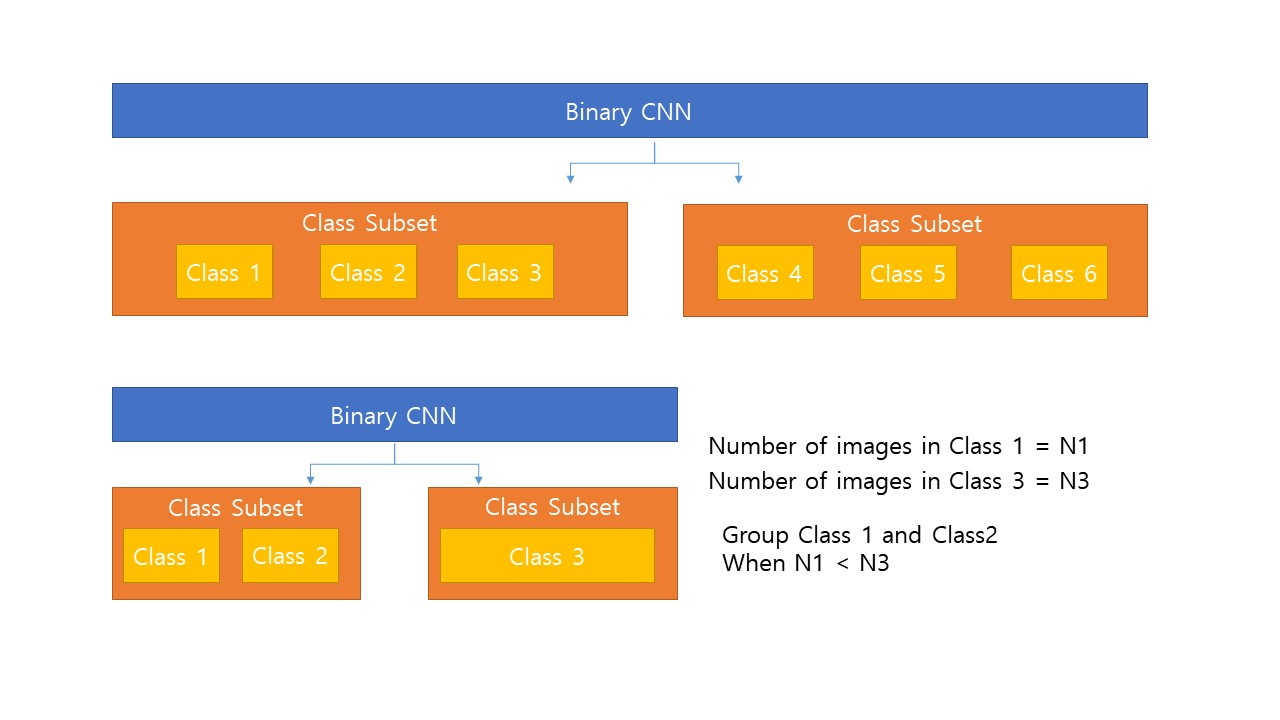}
\caption{Method of choosing K by balancing number of classes. Choose K that balances number of classes. In case of having even number, each subset will have same number of classes. Else, one subset will have one more class. }
\end{figure}

}

\subsubsection{Training and Prediction based on Tournament structure}
After constructing tournament tree, each binary model trained to classify two subset which is divided by criteria K in the tournament tree. For prediction, input is fed into the root model which classifies class set in root node of tournament into the two subset. Iteratively input image is fed into model which classifies the subset which containing class from prediction of previous model into two subset. Final predicted subset which only has one class become prediction of cataract grade.

\subsubsection{Binary deep convolutional neural network}
Assuming that number of class is given N and label is a natural number between 1 and N, Tournament based Ranking CNN consists of N-1 binary deep convolutional neural network that predicts whether label of input image is smaller than k or not where k is a natural number between 1 and N+1. Each binary model has different k. We used pretrained Resnet-101 as binary model.  



Differs from Ranking CNN which used 5 convolution layers and dens layers, our method used very deep convolutional neural network, Resnet\cite{he2016deep}. Reason for using Resnet is that we expected performance gain on accuracy as capacity of model became larger as model goes deeper, and residual network of Resnet prevents over fitting and vanishing gradients. 

Because our cataract dataset consists of relatively small number of image to train whole deep neural network, we used pretrained model with very deep convolutional model to exploit powerful performance of it. Although it is usually used to solve similar problems, according to Shin, Hoo-Chang, et al.\cite{shin2016deep}, transfer learning between very different dataset with medical dataset showed successful result.
 
 Last fully connected layers of pretrained network is replaced to Global Average Pooling layer, and fully connected softmax layer which produces output. Class activation map\cite{zhou2016learning} can be produced by projecting weight of fully connected layer on last convolution feature map. This Class activation map visualize which part of image has importance on prediction. 

\section{Experiment}

In this section, we evaluated performance of Tournament Ranking CNN, Ranking CNN, CNN with linear regression and pretrained Resnet with multi label classification on Cataract grading. We compared our model with Ranking CNN to verify how much better our model is than the previous model by considering unbalanced and vagueness of dataset. Also, we compared CNN with linear regression and pretrained Resnet to evaluate regression model and multi label classification model. We experimented with Intel(R) Core(TM) I7-4790 CPU @ 3.60GHz, NVIDIA TITAN Xp 12GB
, Ubuntu 16.04 (64-bit) and Keras2.

Cataract grade dataset is provided by Korea University hospital. The number of the image per cataract grade is shown in Table 1. Due to distribution of patients to cataract grade, there are very few images on cataract 1 and 6. Scarcity of extreme cataract causes data imbalance between middle cataract grades and grades on both ends.

\begin{table}[h!]
  \begin{center}
    \label{tab:table1}
    \caption{Number of cataract images by cataract grades.}
    \scalebox{1}{
    
    \begin{tabular}{c|c} 
      \textbf{Cataract Grade} & \textbf{Number of image}\\
    
      \hline
      1 & 8\\
      2 & 96\\
      3 & 105\\
      4 & 63\\
      5 & 14\\
      6 & 8\\
      
    \end{tabular}}
    
  \end{center}
\end{table}

We performed 5-fold cross validation over 3 models of tournament Ranking CNN, Ranking CNN, Linear regression with CNN, and Multi label classification with pretrained Resnet. Ranking CNN consists of 5 binary convolutional Neural Network with 5 Convolution layers and 3 Dense layers. Prediction is calculated by aggregating 5 binary outputs of models. Linear regression with CNN consists of 5 Convolution layers , 3 Dense layers and Linear regression layer on top of it. Pretrained Resnet 101 model with Image-net Dataset used to experiment multi label classification model. Top dense layer was replaced with new dense layer consisted of 6 output neurons, each represents one of grades of cataract.
\subsection{Experiment Result}
Table 2 describes result with Accuracy of exact match of cataract grades. In Table 2, Tournament based Ranking CNN with AUC approach(T\_AUC) recorded highest accracy of 68.36 compared to other models. Also, T\_AUC recorded the highest accuracy in grade 1,2,5,6. 
Table 3 describes accuracy of with-in-one-category-off match of cataract grades. T\_AUC showed highest accuracy in grade 1,3,5,6. 

\begin{table}[h!]
  \begin{center}
   \caption{Accuracy(\%) of exact match of cataract grades.}
    \scalebox{0.95}{
    \label{tab:table1}
    \begin{tabular}{c c c c c c c |c} 
      \textbf{       } & \textbf{Grade 1}& \textbf{Grade 2}& \textbf{Grade 3}& \textbf{Grade 4}& \textbf{Grade 5}& \textbf{Grade 6}& \textbf{Average}\\
    
      \hline
      T\_AUC & 50&79&73&54&50&50&68.36\\
      T\_Class & 25&71&69&54&29&25&61.90\\
      T\_Image & 22&71&64&59&43&25&65.98\\
      Ranking CNN & 0&70&60&37&29&0&53.40\\
      CNN\_linear & 0&59&80&40&14&0&57.48\\
      Resnet & 12&69&51&54&43&50&56.12\\
      
    \end{tabular}}
   
  \end{center}
\end{table}

\begin{table}[h!]
  \begin{center}
   \caption{Accuracy(\%) of with-in-one-category-off match.}
    \label{tab:table1}
    \scalebox{0.95}{
    \begin{tabular}{c c c c c c c|c} 
      \textbf{       } & \textbf{Grade 1}& \textbf{Grade 2}& \textbf{Grade 3}& \textbf{Grade 4}& \textbf{Grade 5}& \textbf{Grade 6}& \textbf{Average}\\
    
      \hline
      T\_AUC & 100&98&100&92&93&88&96.93\\
      T\_Class &100&99&100&88&93&50&95.23\\
      T\_Image & 78&98&99&95&93&62&95.91\\
      Ranking CNN &75&96&100&92&64&25&92.85\\
      CNN\_linear & 75&98&100&98&86&12&95.91\\
      Resnet &88&95&99&89&93&50&93.53\\
      
    \end{tabular}}
   
  \end{center}
\end{table}

As the number of images is unbalanced among the grades, accuracy of trained models tends to be higher on the grades which number of images are concentrated. Shown in Table 2, Grade 2 and Grade 3 have higher accuracy compared to other grades. This phenomenon is commonly seen over 6 models, meaning that model is trained to easily predict the grade of input image to be grade which is more common. As a result, error rate of prediction on non common grade increases because model predicts non common grades to be common grades in order to increase accuracy and decrease training loss. This situation is especially severe in Convolution Neural Network with Linear regression model (CNN\_Linear). Specifically in Grade 1 and 6, the model failed to discriminate them. In figure 6, confusion matrix of CNN\_Linear model shows that all images of grade 1 is predicted as grade 2. However, In Tournament based model, this tendency is alleviated because dataset is divided stepwise in training and prediction process. In confusion matrix of 3 of tournament model, Grade with small number of dataset has higher prediction accuracy compared to conventional approaches including Ranking CNN. Especially, Tournament Based Ranking CNN with AUC based model recorded the highest accuracy of 50\% in predicting grade 1 and 6, despite of it's extremely low number of image compared to adjacent grades. By enabling classification of dominated classes and dealing with vagueness by dividing dataset with AUC, T\_AUC was able to achieve highest accuracy. 

{\centering
\begin{figure}
\centering
\begin{tabular}{c c c}

\includegraphics[height=3.4cm]{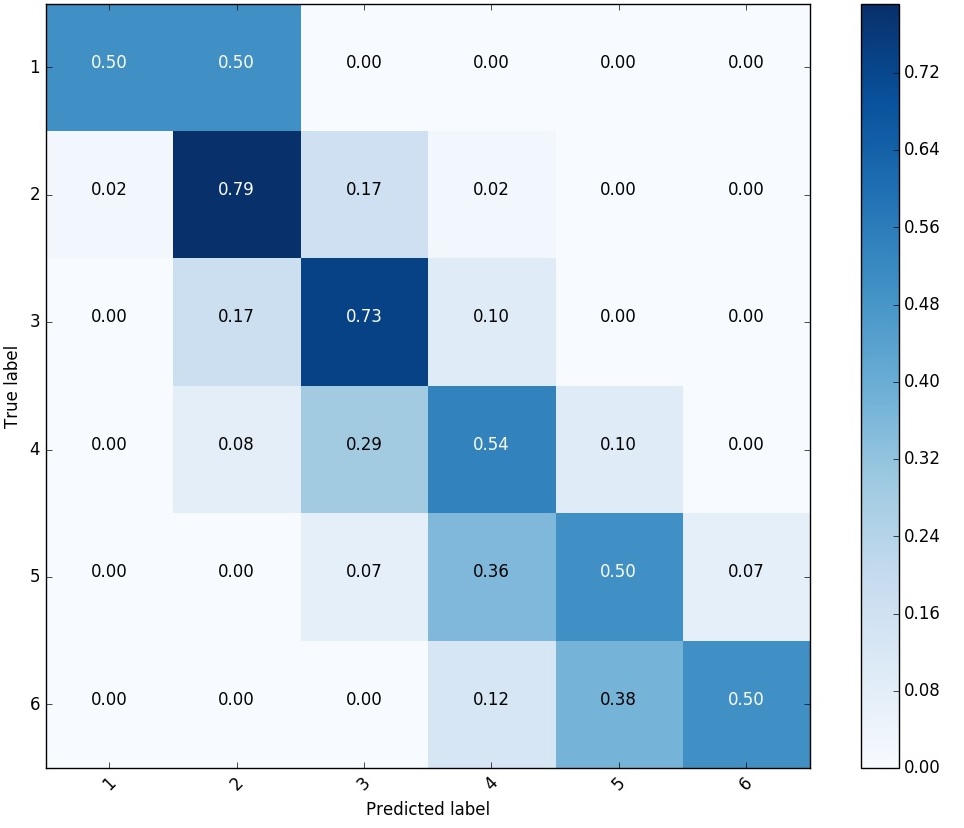}&\includegraphics[height=3.4cm
]{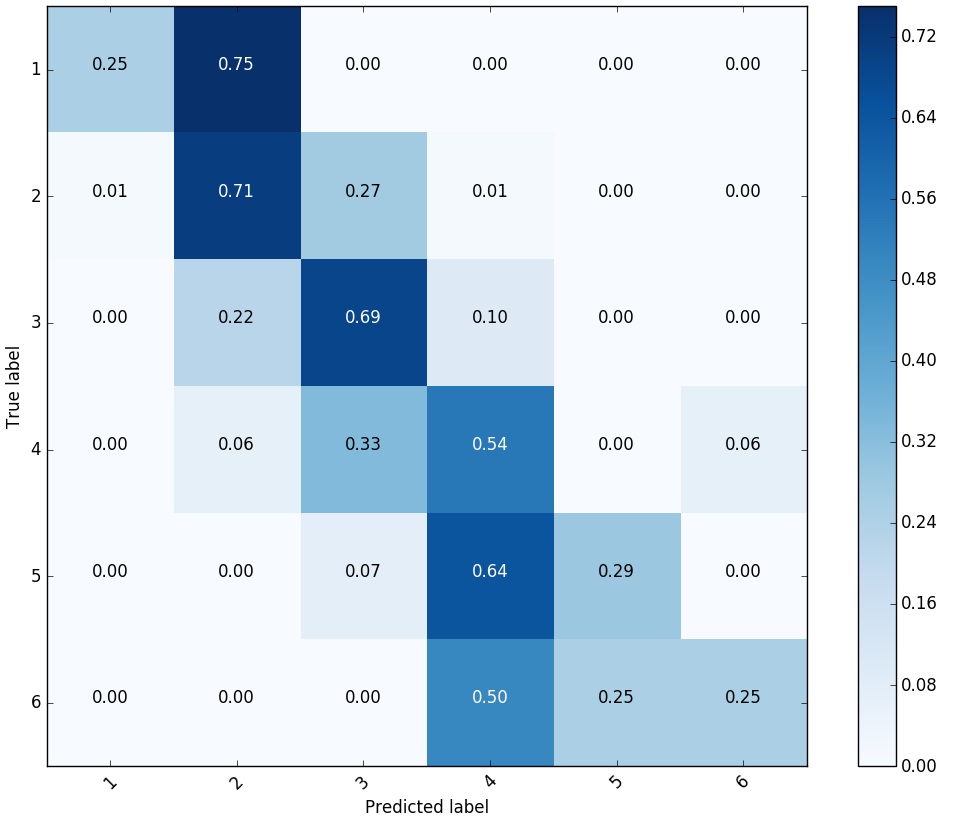}&\includegraphics[height=3.4cm]{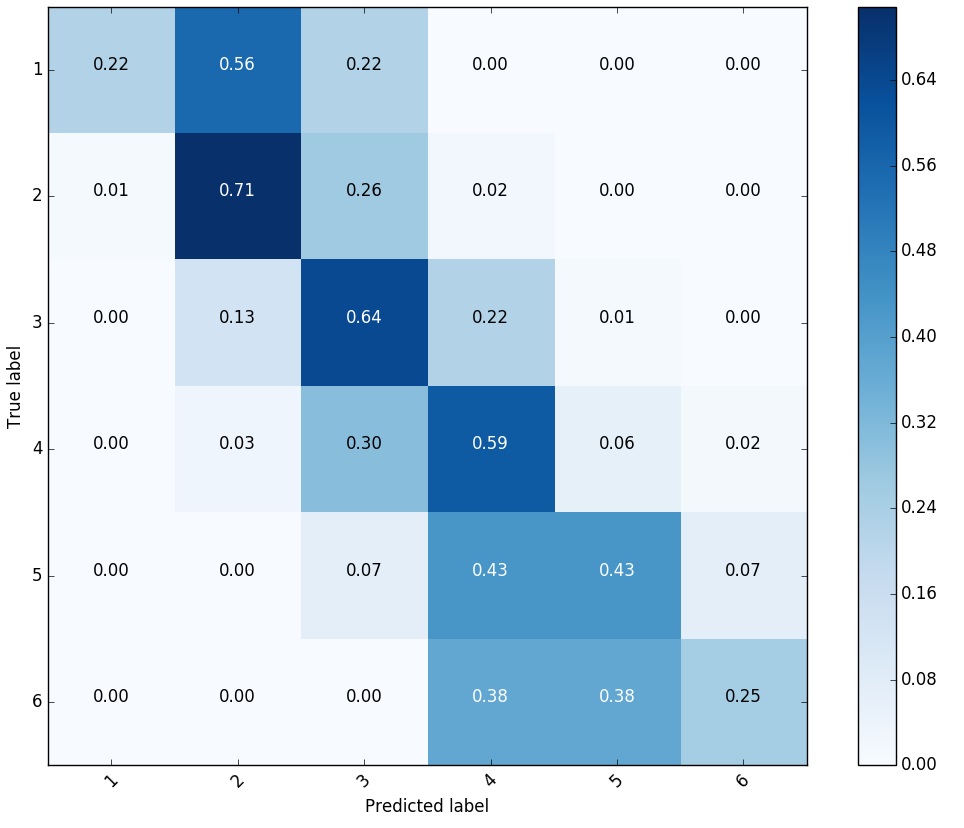}\\ 
T\_AUC & T\_Class & T\_Balance\\
\includegraphics[height=3.4cm]{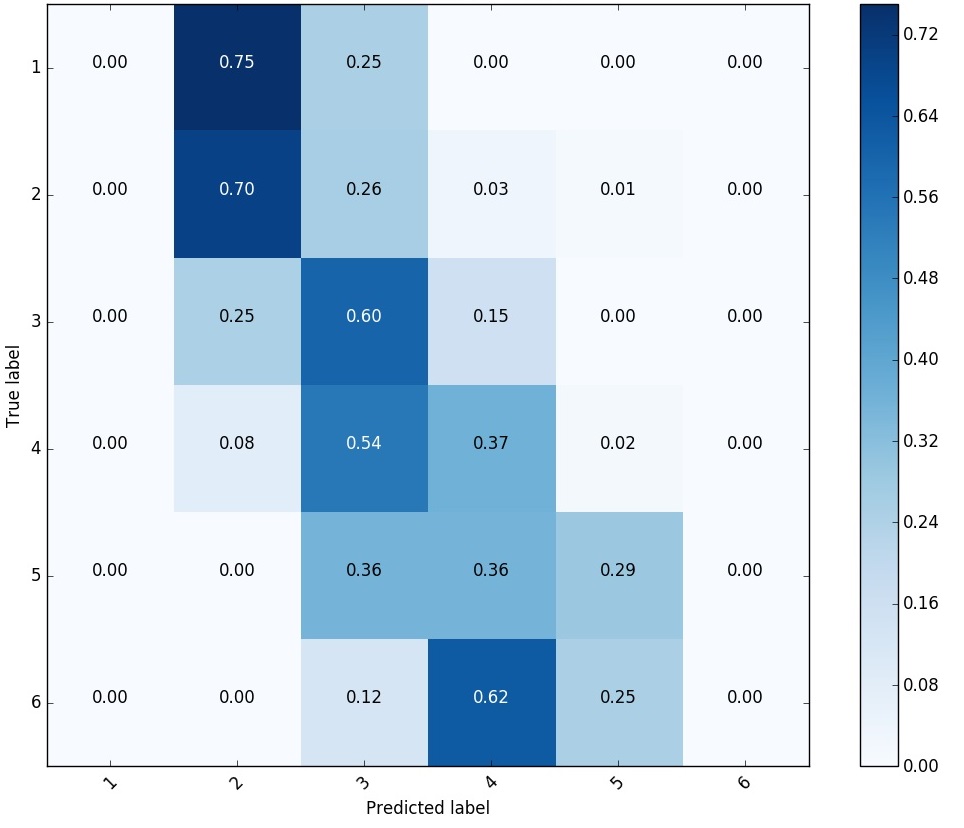}&\includegraphics[height=3.4cm
]{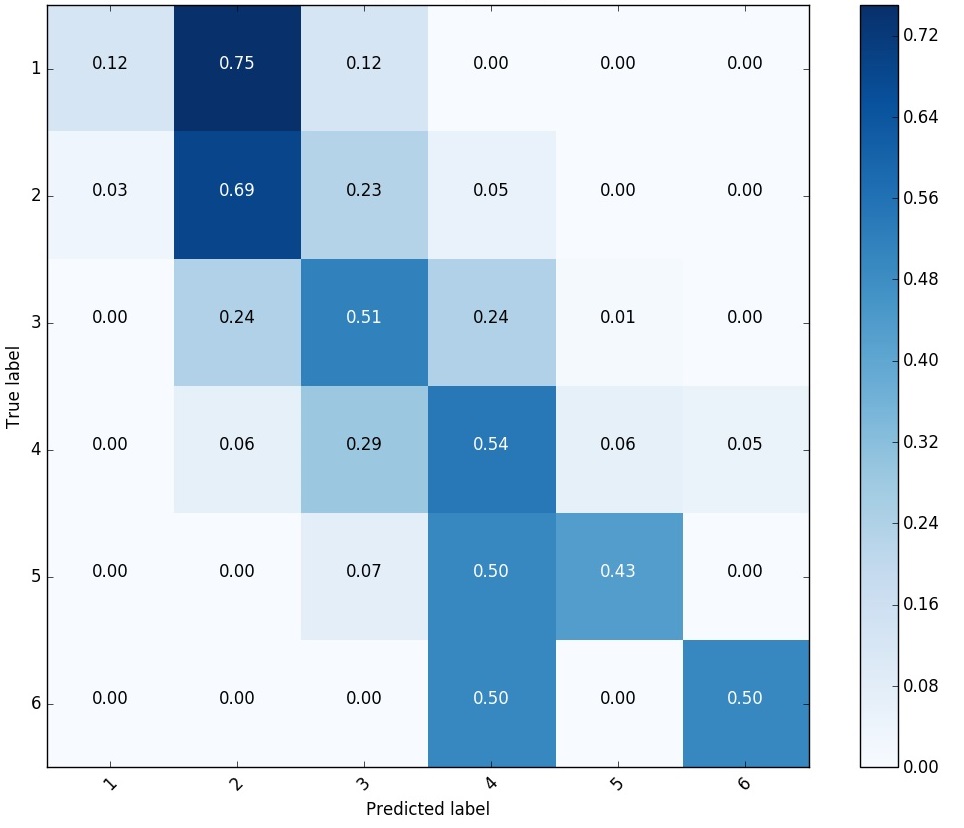}&\includegraphics[height=3.4cm]{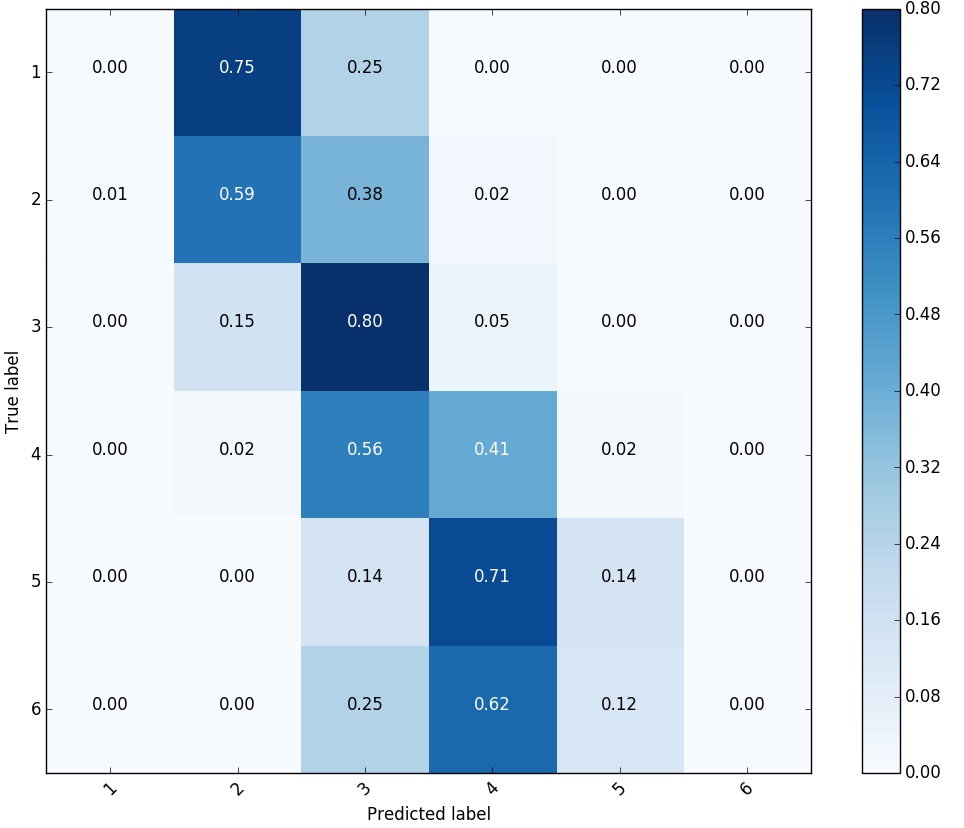}\\ 
RCNN & RESNET & CNN\_Linear\\
\end{tabular}
\caption{Confusion matrix results for cataract grading. }
\end{figure}

}

\subsection{Lens localization with CAM}

To visualize evidence of prediction from model, Our model used Class Activation Map\cite{zhou2016learning} for visualization. Our end of binary pretrained model consists of Convolution layers which generate feature map and Global Average Pooling layer that averages value of feature map, fully connected layer and softmax layer. Class Activation Map\cite{zhou2016learning} of pretrained binary model is achieved by multiplying feature maps from last convolution layer with weight that corresponds to grade $G$ for $K$th feature map. In our case, feature map of last Convolution layer has 512 Channels, and in fully connected layer has 6 nodes for 6 cataract grade. To get $I$th grade of Class Activation Map, $K$th feature map is multiplied with weight of $I$th Node of fully connected layer that connects $K$th feature map. \\
Class Activation Map can be expressed as $\Sigma_k w^I_k f_k$ where $w^I_k$ is weight of $I$th node of fully connected layer that connects $k$th feature map $f_k$\\ Feature map of last Convolution layer, size of 7$\times$7 is bilinear up-sampled to fit original input size of 224$\times$224. Figure 7 shows result of Class activation map of Cataract images. In figure 7, red color indicates colored region has strong evidence of prediction. In contrast blue colored region indicates that region has less impact on prediction. Red colored region is placed in center of eye lens, indicating that center of eye lens effected a lot in predicting cataract grade. In contrast, exterior of eye ball is colored blue, indicating that exterior region of eye ball has very little impact on deciding cataract grade. Although no prior knowledge that crystalline of eye is ROI(Region of Interest) region for classifying cataract grade is given to the model, it clearly acknowledges that center of eye lens where crystalline is located has critical clue for predicting cataract grade. 

{\centering
\begin{figure}
\centering
\begin{tabular}{c c c}
\includegraphics[height=3.2cm]{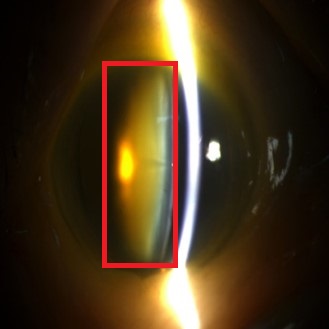}&\includegraphics[height=3.2cm]{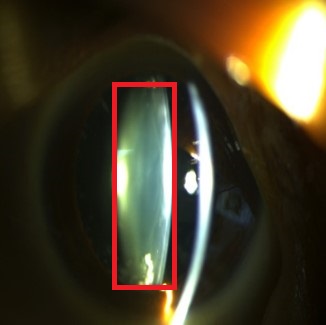}&\includegraphics[height=3.2cm]{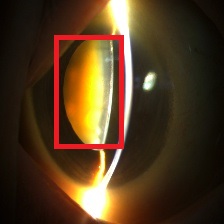}\\
\includegraphics[height=3.2cm]{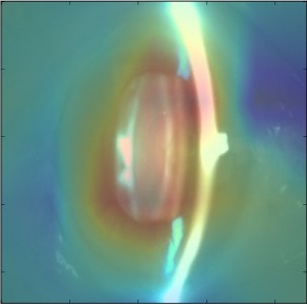} &
\includegraphics[height=3.2cm]{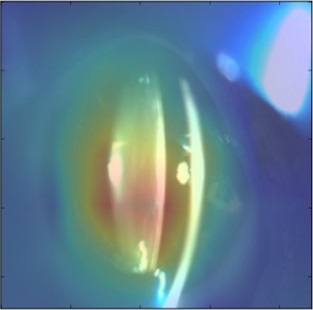} &
\includegraphics[height=3.2cm]{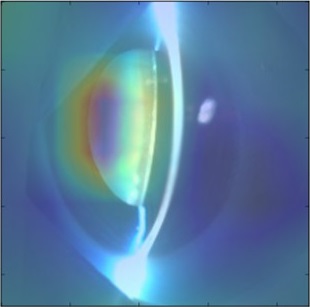}
\end{tabular}
\caption{ROI of original Slit-lamp images on top side, and Class Activation Map images identifying ROI bottom side. Red regions indicates strong evidence in Class Activation Map.}
\end{figure}

}

\section{Conclusion}

Imbalance number of images among classes hinder the model to train the classes which has small number of data compared to the others. With conventional approach, it is natural that having low accuracy on the class with small data. This is because model tries to ignore small classes and get higher accuracy in a large classes in order to get higher overall accuracy. In cataract dataset, number of the severe cataract grade image is very small due to scarcity of patients having it. As a result, accuracy of severe grades, class over grade 5 was very low even importance of classifying it is very high in a sense of preventing cataract. 

Ambiguity of the boundary between cataract grades is also one of the reason that degrades accuracy. As opacity of crystalline lens is gradually increase with grades, it is difficult to set clear criteria for dividing each classes.

Also, as Ranking CNN\cite{chen2017using} pointed out, cataract grading approach with regression models and multi label classification models degrades performance because regression models over-simplifies the opaque patterns of cataract into linear forms and multi label classification ignores ordinal properties of cataract grades. 

For these problems, We proposed Tournament Based Ranking CNN. By dividing training and test set in two sets iteratively, model alleviates being learned to be biased toward a class having large number of data. Also, by dividing classes into two set in a way that recorded highest AUC from classification of two divided set, model was able to achieve higher accuracy despite of ambiguous boundary of cataract grades. Moreover, our model was able to exploit ordinal property of cataract dataset by using multiple binary models with tournament structure. 

Finally, CAM layer successfully visualized which part of image was based on prediction of our proposed model. Unlike previous works used ROI detection algorithms before prediction, our model was able to capture ROI of cataract image without any prior knowledges.

Thus we expect our approaches can be applied more generalized problem with various area, which suffers from imbalance of data and vagueness with ordinal dataset.



\bibliographystyle{splncs}
\bibliography{egbib}

\begin{thebibliography}{10}

\bibitem{nayak2013automated}
Nayak, J.:
\newblock Automated classification of normal, cataract and post cataract
  optical eye images using svm classifier.
\newblock In: Proceedings of the World Congress on Engineering and Computer
  Science. Volume~1. (2013)  23--25

\bibitem{huang2009computer}
Huang, W., Li, H., Chan, K.L., Lim, J.H., Liu, J., Wong, T.Y.:
\newblock A computer-aided diagnosis system of nuclear cataract via ranking.
\newblock In: International Conference on Medical Image Computing and
  Computer-Assisted Intervention, Springer (2009)  803--810

\bibitem{li2010computer}
Li, H., Lim, J.H., Liu, J., Mitchell, P., Tan, A.G., Wang, J.J., Wong, T.Y.:
\newblock A computer-aided diagnosis system of nuclear cataract.
\newblock IEEE Transactions on Biomedical Engineering \textbf{57} (2010)
  1690--1698

\bibitem{xu2013automatic}
Xu, Y., Gao, X., Lin, S., Wong, D.W.K., Liu, J., Xu, D., Cheng, C.Y., Cheung,
  C.Y., Wong, T.Y.:
\newblock Automatic grading of nuclear cataracts from slit-lamp lens images
  using group sparsity regression.
\newblock In: International Conference on Medical Image Computing and
  Computer-Assisted Intervention, Springer (2013)  468--475

\bibitem{liu2017localization}
Liu, X., Jiang, J., Zhang, K., Long, E., Cui, J., Zhu, M., An, Y., Zhang, J.,
  Liu, Z., Lin, Z.,  et~al.:
\newblock Localization and diagnosis framework for pediatric cataracts based on
  slit-lamp images using deep features of a convolutional neural network.
\newblock PloS one \textbf{12} (2017)  e0168606

\bibitem{gao2015automatic}
Gao, X., Lin, S., Wong, T.Y.:
\newblock Automatic feature learning to grade nuclear cataracts based on deep
  learning.
\newblock IEEE Transactions on Biomedical Engineering \textbf{62} (2015)
  2693--2701

\bibitem{yang2016exploiting}
Yang, J.J., Li, J., Shen, R., Zeng, Y., He, J., Bi, J., Li, Y., Zhang, Q.,
  Peng, L., Wang, Q.:
\newblock Exploiting ensemble learning for automatic cataract detection and
  grading.
\newblock Computer methods and programs in biomedicine \textbf{124} (2016)
  45--57

\bibitem{chen2017using}
Chen, S., Zhang, C., Dong, M., Le, J., Rao, M.:
\newblock Using ranking-cnn for age estimation.
\newblock In: The IEEE Conference on Computer Vision and Pattern Recognition
  (CVPR). (2017)

\bibitem{jun20182sranking}
Jun, T.J., Kim, D., Nguyen, H.M., Kim, D., Eom, Y.:
\newblock 2sranking-cnn: A 2-stage ranking-cnn for diagnosis of glaucoma from
  fundus images using cam-extracted roi as an intermediate input.
\newblock arXiv preprint arXiv:1805.05727 (2018)

\bibitem{he2016deep}
He, K., Zhang, X., Ren, S., Sun, J.:
\newblock Deep residual learning for image recognition.
\newblock In: Proceedings of the IEEE conference on computer vision and pattern
  recognition. (2016)  770--778

\bibitem{shin2016deep}
Shin, H.C., Roth, H.R., Gao, M., Lu, L., Xu, Z., Nogues, I., Yao, J., Mollura,
  D., Summers, R.M.:
\newblock Deep convolutional neural networks for computer-aided detection: Cnn
  architectures, dataset characteristics and transfer learning.
\newblock IEEE transactions on medical imaging \textbf{35} (2016)  1285--1298

\bibitem{zhou2016learning}
Zhou, B., Khosla, A., Lapedriza, A., Oliva, A., Torralba, A.:
\newblock Learning deep features for discriminative localization.
\newblock In: Computer Vision and Pattern Recognition (CVPR), 2016 IEEE
  Conference on, IEEE (2016)  2921--2929

\end{thebibliography}

\end{document}